\documentclass[sigconf]{acmart}

\usepackage{booktabs} 
\usepackage{bm}
\usepackage{comment}
\usepackage{soul}
\usepackage{caption}
\usepackage{float}
\usepackage{amsfonts}
\usepackage[normalem]{ulem}
\usepackage[]{algorithm2e}
\usepackage{todonotes}
\usepackage{multirow}
\usepackage{hyperref}
\usepackage{subcaption,siunitx,booktabs}

\copyrightyear{2018}
\acmYear{2018} 
\setcopyright{iw3c2w3}
\acmConference[WWW 2018]{The 2018 Web Conference}{April 23--27, 2018}{Lyon, France}
\acmBooktitle{WWW 2018: The 2018 Web Conference, April 23--27, 2018, Lyon, France}
\acmPrice{}
\acmDOI{10.1145/3178876.3186040}
\acmISBN{978-1-4503-5639-8/18/04}
\begin{document}
\title{Field-weighted Factorization Machines for Click-Through Rate Prediction in Display Advertising}

\author{Junwei~Pan}
\affiliation{Yahoo Research}
\email{jwpan@oath.com}
\author{Jian~Xu}
\affiliation{TouchPal Inc.}
\email{nayobux@gmail.com}
\author{Alfonso~Lobos~Ruiz}
\affiliation{UC Berkeley}
\email{alobos@berkeley.edu}
\author{Wenliang~Zhao}
\affiliation{Yahoo Research}
\email{wenliangz@oath.com}
\author{Shengjun~Pan}
\affiliation{Yahoo Research}
\email{alanpan@oath.com}
\author{Yu~Sun}
\affiliation{LinkedIn Corporation}
\email{ysun1@linkedin.com}
\author{Quan~Lu}
\affiliation{Ablibaba Group}
\email{quan.lu@gmail.com}

\renewcommand{\shortauthors}{J. Pan et al.}

\begin{abstract}
Click-through rate (CTR) prediction is a critical task in online display advertising. The data involved in CTR prediction are typically multi-field categorical data, i.e., every feature is categorical and belongs to one and only one field. One of the interesting characteristics of such data is that features from one field often interact differently with features from different other fields. Recently, Field-aware Factorization Machines (FFMs) have been among the best performing models for CTR prediction by explicitly modeling such difference. However, the number of parameters in FFMs is in the order of feature number times field number, which is unacceptable in the real-world production systems. In this paper, we propose Field-weighted Factorization Machines (FwFMs) to model the different feature interactions between different fields in a much more memory-efficient way. Our experimental evaluations show that FwFMs can achieve competitive prediction performance with only as few as $4\%$ parameters of FFMs. When using the same number of parameters, FwFMs can bring $0.92\%$ and $0.47\%$ AUC lift over FFMs on two real CTR prediction data sets.
\end{abstract}

%
%

\ccsdesc[500]{Computing methodologies~Factorization methods}
\ccsdesc[500]{Information systems~Computational advertising}
\ccsdesc[300]{Theory of computation~Computational advertising theory}

\keywords{Display Advertising; CTR Prediction; Factorization Machines}

\maketitle

\section{Introduction}\label{sec:intro}
Online display advertising is a multi-billion dollar business nowadays, with an annual revenue of 31.7 billion US dollars in fiscal year 2016, up 29\% from fiscal year 2015~\cite{iab-report}. One of the core problems display advertising strives to solve is to deliver the right ads to the right people, in the right context, at the right time. Accurately predicting the click-through rate (CTR) is crucial to solve this problem and it has attracted much research attention in the past few years~\cite{chapelle2015simple, mcmahan2013ad, richardson2007predicting}.

The data involved in CTR prediction are typically \emph{multi-field categorical data}~\cite{zhang2016deep} which are also quite ubiquitous in many applications besides display advertising, such as recommender systems~\cite{rendle2010factorization}. Such data possess the following properties. First, all the features are categorical and are very sparse since many of them are identifiers. Therefore, the total number of features can easily reach millions or tens of millions. Second, every feature belongs to one and only one field and there can be tens to hundreds of fields. Table 1 is an example of a real-world multi-field categorical data set used for CTR prediction. 

\begin{table}
\centering
\begin{tabular}{ | c | c | c | c | c | } 
    \hline
    \texttt{\textbf{CLICK}} & \texttt{\textbf{User\_ID}} & \texttt{\textbf{GENDER}} & \texttt{\textbf{ADVERTISER}} & \texttt{\textbf{PUBLISHER}}  \\
    \hline
    1 & 29127394 & \texttt{Male} & \texttt{Nike} & \texttt{news.yahoo.com} \\
    \hline
    -1 & 89283132 & \texttt{Female} & \texttt{Walmart} & \texttt{techcrunch.com} \\
    \hline
    -1 & 91213212 & \texttt{Male} & \texttt{Gucci} & \texttt{nba.com} \\
    \hline
    -1 & 71620391 & \texttt{Female} & \texttt{Uber} & \texttt{tripadviser.com} \\
    \hline
    1 & 39102740 & \texttt{Male} & \texttt{Adidas} & \texttt{mlb.com} \\
    \hline
 	\end{tabular}
    \caption{An example of multi-field categorical data for CTR prediction. Each row is an ad impression. Column \texttt{CLICK} is the label indicating whether there is a click associated with this impression. Each of the rest columns is a field. Features are all categorical, e.g., \texttt{Male}, \texttt{Female}, \texttt{Nike}, \texttt{Walmart}, and each of them belongs to one and only one field, e.g., \texttt{Male} belongs to field \texttt{GENDER}, and \texttt{Nike} belongs to field \texttt{ADVERTISER}.}
\label{table:artificial}
\end{table}

The properties of multi-field categorical data pose several unique challenges to building effective machine learning models for CTR prediction:

\begin{enumerate}
	\item \textbf{Feature interactions are prevalent and need to be specifically modeled}~\cite{chapelle2015simple,chang2010training}. Feature conjunctions usually associate with the labels differently from individual features do. For example, the CTR of \texttt{Nike}'s ads shown on \texttt{nba.com} is usually much higher than the average CTR of \texttt{Nike}'s ads or the average CTR of ads shown on \texttt{nba.com}. This phenomenon is usually referred to as \emph{feature interaction} in the literature~\cite{juan2017field}. To avoid confusion and simplify discussion, in the rest of the paper, we unify and abuse the terminology \emph{feature interaction strength} to represent the level of association between feature conjunctions and labels.
	\item \textbf{Features from one field often interact differently with features from different other fields}. For instance, we have observed that features from field \texttt{GENDER} usually have strong interactions with features from field \texttt{ADVERTISER} while their interactions with features from field \texttt{DEVICE\_TYPE} are relatively weak. This might be attributed to the fact that users with a specific gender are more biased towards the ads they are viewing than towards the type of device they are using.
	\item \textbf{Potentially high model complexity needs to be taken care of}~\cite{juan2017field}. The model parameters such as weights and embedding vectors need to be stored in memory in real-world production systems to enable real-time ad serving. As there are typically millions of features in practice, the model complexity needs to be carefully designed and tuned to fit the model into memory.
\end{enumerate}

To resolve part of these challenges, researchers have built several solutions. Factorization Machines (FMs)~\cite{rendle2010factorization,rendle2012factorization} and Field-aware Factorization Machines (FFMs)~\cite{juan2016field,juan2017field} are among the most successful ones. FMs tackle the first challenge by modeling the effect of pairwise feature interactions as the dot product of two embedding vectors. However, the field information is not leveraged in FMs at all. Recently, FFMs have been among the best performing models for CTR prediction and won two competitions hosted by Criteo and Avazu~\cite{juan2016field}. FFMs learn different embedding vectors for each feature when the feature interacts with features from different other fields. In this way, the second challenge is explicitly tackled. However, the number of parameters in FFMs is in the order of feature number times field number, which can easily reach tens of millions or even more. This is unacceptable in real-world production systems. In this paper, we introduce Field-weighted Factorization Machines (FwFMs) to resolve all these challenges simultaneously. The main contributions of this paper can be summarized as follows:

\begin{enumerate}
	\item Empirically we show that the average interaction strength of feature pairs from one field pair tends to be quite different from that of other field pairs. In other words, different field pairs have significantly different levels of association with the labels (i.e. clicks in CTR prediction). Following the same convention, we call this \emph{field pair interactions}. 
    \item Based on the above observation, we propose Field-weighted Factorization Machines (FwFMs). By introducing and learning a field pair weight matrix, FwFMs can effectively capture the heterogeneity of field pair interactions. Moreover, parameters in FwFMs are magnitudes fewer than those in FFMs, which makes FwFMs a preferable choice in real-world production systems.
    \item FwFMs are further augmented by replacing the binary representations of the linear terms with embedding vector representations. This novel treatment can effectively help avoid over-fittings and enhance prediction performance.
    \item We conduct comprehensive experiments on two real-world CTR prediction data sets to evaluate the performance of FwFMs against existing models. The results show that FwFMs can achieve competitive prediction performance with only as few as $4\%$ parameters of FFMs. When using the same number of parameters, FwFMs outperform FFMs by up to $0.9\%$ AUC lift. 
\end{enumerate}

The rest of the paper is organized as follows. Section~\ref{sec:pre} provides the preliminaries of existing CTR prediction models that handle multi-field categorical data. In Section~\ref{sec:interaction}, we show that the interaction strengths of different field pairs are quite different, followed by detailed discussion of the proposed model in Section~\ref{sec:fwFM}. Our experimental evaluation results are presented in Section~\ref{sec:exp}. In Section~\ref{sec:analysis}, we show that FwFMs learn the field pair interaction strengths even better than FFMs. Section~\ref{sec:related} and Section~\ref{sec:conclusion} discusses the related work and concludes the paper respectively.

\section{Preliminaries}\label{sec:pre}




Logistic Regression (LR) is probably the most widely used machine learning model on multi-field categorical data for CTR prediction~\cite{chapelle2015simple,richardson2007predicting}. Suppose there are $m$ unique features $\{f_1,\cdots,f_{m}\}$ and $n$ different fields $\{F_1,\cdots,F_{n}\}$. Since each feature belongs to one and only one field, to simplify notation, we use index $i$ to represent feature $f_i$ and $F(i)$ to represent the field $f_i$ belongs to. Given a data set $\bm{S} = \{y^{(s)},\bm{x}^{(s)}\}$, where $y^{(s)}\in\{1,-1\}$ is the label indicating click or non-click and $\bm{x}^{(s)} \in \{0,1\}^m$ is the feature vector in which $x_i^{(s)} = 1$ if feature $i$ is active for this instance otherwise $x_i^{(s)}=0$, the LR model parameters $\bm{w}$ are estimated by minimizing the following regularized loss function:
\begin{equation}
	\min_{\bm{w}}\;\lambda \|\bm{w}\|_2^2 + \sum_{s=1}^{|S|}\log(1+\exp(-y^{(s)} \Phi_{LR}(\bm{w}, \bm{x}^{(s)})))
\end{equation}
where $\lambda$ is the regularization parameter, and
\begin{equation}
\Phi_{LR}(\bm{w}, \bm{x}) = w_0 + \sum_{i=1}^m x_i w_i
\end{equation}
is a linear combination of individual features.

However, linear models are not sufficient for tasks such as CTR prediction in which feature interactions are crucial~\cite{chapelle2015simple}. A general way to address this problem is to add feature conjunctions. It has been shown that Degree-2 Polynomial (Poly2) models can effectively capture the effect of feature interactions\cite{chang2010training}. Mathematically, in the loss function of equation (1), Poly2 models consider replacing $\Phi_{LR}$ with
\begin{equation}
\Phi_{Poly2}(\bm{w}, \bm{x}) = w_0 + \sum_{i=1}^m x_i w_i + \sum_{i=1}^{m}\sum_{j=i+1}^m x_{i}x_{j}w_{h(i,j)}
\end{equation}
where $h(i,j)$ is a function hashing $i$ and $j$ into a natural number in hashing space $H$ to reduce the number of parameters. Otherwise the number of parameters in the model would be in the order of $O(m^2)$.

Factorization Machines(FMs) learn an embedding vector $\bm{v}_i \in \mathbb{R}^K$ for each feature, where $K$ is a hyper-parameter and is usually a small integer, e.g., 10. FMs model the interaction between two features $i$ and $j$ as the dot product of their corresponding embedding vectors $\bm{v}_i$, $\bm{v}_j$:

\begin{equation}
\Phi_{FMs}\left((\bm{w},\bm{v}),\bm{x} \right)=  w_0 + \sum_{i=1}^m x_i w_i +\sum_{i=1}^{m}\sum_{j=i+1}^m x_{i}x_{j} \langle \bm{v}_{i}, \bm{v}_{j}\rangle
\end{equation}

FMs usually outperform Poly2 models in applications involving sparse data such as CTR prediction. The reason is that FMs can always learn some meaningful embedding vector for each feature as long as the feature itself appears enough times in the data, which makes the dot product a good estimate of the interaction effect of two features even if they have never or seldom occurred together in the data. However, FMs neglect the fact that a feature might behave differently when it interacts with features from different other fields. Field-aware Factorization Machines (FFMs) model such difference explicitly by learning $n-1$ embedding vectors for each feature, say $i$, and only using the corresponding one $\bm{v}_{i, F(j)}$ to interact with another feature $j$ from field $F(j)$:
\begin{equation}
	\Phi_{FFMs}((\bm{w}, \bm{v}), \bm{x})= w_0 + \sum_{i=1}^m x_i w_i + \sum_{i=1}^{m}\sum_{j=i+1}^m x_{i}x_{j} \langle \bm{v}_{i, F(j)}, \bm{v}_{j, F(i)}\rangle 
\end{equation}

Although FFMs have got significant performance improvement over FMs, their number of parameters is in the order of $O(mnK)$. The huge number of parameters of FFMs is undesirable in the real-world production systems. Therefore, it is very appealing to design alternative approaches that are competitive and more memory-efficient.

\section{Interaction Strengths of Field Pairs}\label{sec:interaction}


In multi-field categorical data, every feature belongs to one and only one field. We are particularly interested in whether the strength of interactions are different at the field level. In other words, whether the average interaction strength between all feature pairs from a field pair is different from that of other field pairs.

For example, in the CTR prediction data, features from field \texttt{ADVERTISER} usually have strong interaction with features from field \texttt{PUBLISHER} since advertisers usually target a group of people with specific interest and the audience of publishers are naturally grouped by interests. On the other hand, features from field \texttt{HOUR\_OF\_DAY} tend to have little interaction with features from field \texttt{DAY\_OF\_WEEK}, which is not hard to understand since by intuition their interactions reveal little about the clicks. 

To validate the heterogeneity of the field pair interactions, we use mutual information~\cite{cover2012elements} between a field pair $(F_k, F_l)$ and label variable $Y$ to quantify the interaction strength of the field pair:
\begin{equation}\label{eq:MI}
	MI((F_k, F_l), Y) = \sum_{(i,j) \in (F_k, F_l)} \sum_{y \in Y} p((i, j), y) log \frac{p((i, j), y)}{p(i,j)p(y)}
\end{equation}


\begin{figure}[!t]
\centering
    \includegraphics[width=\columnwidth]{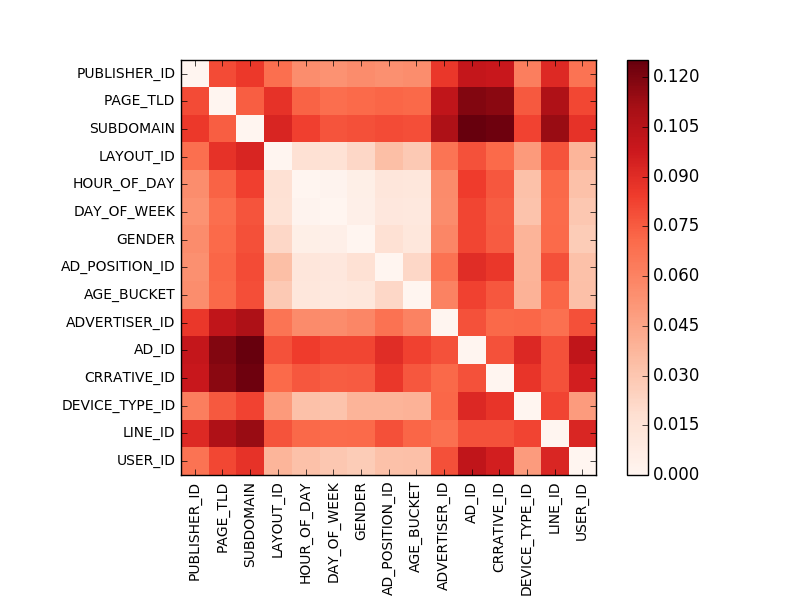}
    \caption{Heat map of mutual information between each field pair and the label.}
    \label{fig:mi}
\end{figure}

Figure~\ref{fig:mi} is a visualization of the mutual information between each field pair and the label, computed from Oath CTR data. Unsurprisingly, the interaction strengths of different field pairs are quite different. Some field pairs have very strong interactions, such as (\texttt{AD\_ID, SUBDOMAIN}), (\texttt{CREATIVE\_ID}, \texttt{PAGE\_TLD}) while some other field pairs have very weak interactions, such as (\texttt{LAYOUT\_ID}, \texttt{GENDER}), (\texttt{DAY\_OF\_WEEK}, \texttt{AD\_POSITION\_ID}). The meanings of the these fields are explained in Section~\ref{subsec:datasets}

Although the analysis result is not surprising, we note that none of the existing models takes this field level interaction heterogeneity into consideration. This motivated us to build an effective machine learning model to capture the different interaction strengths of different field pairs.

\section{Field-weighted Factorization Machines (FwFMs)}\label{sec:fwFM}


We propose to explicitly model the different interaction strengths of different field pairs. More specifically, the interaction of a feature pair $i$ and $j$ in our proposed approach is modeled as

\[x_ix_j\langle \bm{v}_i, \bm{v}_j\rangle r_{F(i), F(j)}\]

\noindent where  $\bm{v}_i, \bm{v}_j$ are the embedding vectors of $i$ and $j$, $F(i), F(j)$ are the fields of feature $i$ and $j$, respectively, and $r_{F(i), F(j)} \in \mathbb{R}$ is a weight to model the interaction strength between field $F(i)$ and $F(j)$. We refer to the resulting model as the Field-weighted Factorization Machines(FwFMs):
\begin{equation}\label{eq:fwfm}
	\Phi_{FwFMs}((\bm{w},\bm{v}), \bm{x}) = w_0 + \sum_{i=1}^m x_i w_i +  \sum_{i=1}^m\sum_{j=i+1}^m x_{i} x_{j} \langle \bm{v}_i, \bm{v}_j \rangle r_{F(i), F(j)}
\end{equation}

FwFMs are extensions of FMs in the sense that we use additional weight $r_{F(i), F(j)}$ to explicitly capture different interaction strengths of different field pairs. FFMs can model this implicitly since they learn several embedding vectors for each feature $i$, each one $\bm{v}_{i, F_k}$ corresponds to one of other fields $F_k \neq F(i)$, to model its different interaction with features from different fields. However, the model complexity of FFMs is significantly higher than that of FMs and FwFMs.

\begin{table}
\centering
	\begin{tabular}{| c | c |}
    \hline
    Model & Number of Parameters \\
    \hline
    LR & $m$ \\
    \hline
    Poly2 & $m + H$ \\
    \hline
    FMs & $m + mK$ \\
    \hline
    FFMs & $ m + m(n-1)K$ \\
    \hline
    FwFMs & $m + mK + \frac{n(n-1)}{2}$ \\
    \hline
	\end{tabular}
    \caption{A summary of model complexities (ignoring the bias term $w_0$). $m$ and $n$ are feature number and field number respectively, $K$ is the embedding vector dimension, and $H$ is the hashing space size when hashing tricks are used for Poly2 models.}
    \label{table:model_complexity}
\end{table}

\subsection{Model Complexity}
\label{subsec:model_complexity}

The number of parameters in FMs is $m + mK$, where $m$ accounts for the weights for each feature in the linear part $\{w_i|i=1,...,m\}$ and $mK$ accounts for the embedding vectors for all the features $\{\bm{v}_i|i=1,...,m\}$. FwFMs use $n(n-1)/2$ additional parameters $\{r_{F_k, F_l} | k, l = 1,...,n\}$ for each field pair so that the total number of parameters of FwFMs is $m + mK + n(n-1)/2$. For FFMs, the number of parameters is $m + m(n-1)K$ since each feature has $n-1$ embedding vectors. Given that usually  $n\ll m$, the parameter number of FwFMs is comparable with that of FMs and significantly less than that of FFMs. In Table~\ref{table:model_complexity} we compare the model complexity of all models mentioned so far.

\subsection{Linear Terms}
\label{subsec:linear-terms}
In the linear terms $\sum_{i=1}^m x_i w_i$ of equation \eqref{eq:fwfm}, we learn a weight $w_i$ for each feature to model its effect with the label, using binary variable $x_i$ to represent feature $i$. However, the embedding vectors $\bm{v}_i$ learned in the interaction terms should capture more information about feature $i$, therefore we propose to use $x_i\bm{v}_i$ to represent each feature in the linear terms as well. 

We can learn one linear weight vector $\bm{w}_i$ for each feature and the linear terms become:

\begin{equation}
	\sum_{i=1}^m x_i \langle \bm{v}_i, \bm{w}_i \rangle
\end{equation}

There are $mK$ parameters in the feature-wise linear weight vectors, and the total number of parameters is $2mK + \frac{n(n-1)}{2}$. Alternatively, we can learn one linear weight vector $\bm{w}_{F(i)}$ for each field and all features from the same field $F(i)$ use the same linear weight vector. Then the linear terms can be formulated as:

\begin{equation}
	\sum_{i=1}^m x_i \langle \bm{v}_i, \bm{w}_{F(i)} \rangle
\end{equation}

The parameter number of these kind of FwFMs is $nK + mK + \frac{n(n-1)}{2}$, which is almost the same as FwFMs with original linear weights since both $K$ and $n$ are usually in the order of tens.

In the rest of the paper, we denote FwFMs with original linear weights as FwFMs\_LW, denote FwFMs with feature-wise linear weight vectors as FwFMs\_FeLV, denote FwFMs with field-wise linear weight vectors as FwFMs\_FiLV.

\section{Experiments}\label{sec:exp}
In this section we present our experimental evaluations results. We will first describe the data sets and implementation details in Section~\ref{subsec:datasets} and ~\ref{subsec:implementation} respectively. In Section~\ref{subsec:performance_comparison} we compare FwFMs\_LW, i.e., FwFMs using original linear weights, with LR, Poly2, FMs and FFMs. Then we further investigate the performance of FwFMs\_LW and FFMs when they use the same number of parameters in Section~\ref{subsec:same_parameter_number}. The enhancement brought by our novel linear term treatment is discussed in Section~\ref{subsec:performance_linear_terms}. Finally we show model hyper-parameter tuning details in Section~\ref{subsec:parameter_tuning}.

\subsection{Data sets}
\label{subsec:datasets}
We use the following two data sets in our experiments.
\begin{enumerate}
	\item Criteo CTR data set: This is the data set used for the Criteo Display Advertising Challenge~\cite{criteo-display-ad-challenge}. We split the data into training, validation and test sets randomly by 60\%:20\%:20\%.
    \item Oath CTR data set: We use two-week display advertising click log from our ad serving system as the training set and the log of next day and the day after next day as validation and test set respectively. 
\end{enumerate}

The Criteo data set is already label balanced. For the Oath CTR data set, the ratio of positive samples (clicks) is the overall CTR, which is typically smaller than $1\%$. We downsample the negative examples so that the positive and negative samples are more balanced. The downsampling is not done for validation and test sets, since the evaluation should be applied to data sets reflecting the actual traffic distribution.

There are 26 anonymous categorical fields in Criteo data set. The Oath data set consists of 15 fields, which can be categorized into 4 groups: user side fields such as \texttt{GENDER}, \texttt{AGE\_BUCKET}, \texttt{USER\_ID}, publisher side fields such as \texttt{PAGE\_TLD}, \texttt{PUBLISHER\_ID}, \texttt{SUBDOMAIN},  advertiser side fields such as \texttt{ADVERTISER\_ID}, \texttt{AD\_ID}, \texttt{CREATIVE\_ID}, \texttt{LAYOUT\_ID} , \texttt{LINE\_ID} and context side fields such as \texttt{HOUR\_OF\_DAY}, \texttt{DAY\_OF\_WEEK}, \texttt{AD\_POSITION\_ID} and \texttt{DEVICE\_TYPE\_ID}. The meanings of most fields are quite straightforward and we only explain some of them: \texttt{PAGE\_TLD} denotes the top level domain of a web page and \texttt{SUBDOMAIN} denotes the sub domain of a web page. \texttt{CREATIVE\_ID} denotes the identifier of a creative, while \texttt{AD\_ID} identifies an ad that encapsulates a creative; the same creative may be assigned to different ads. \texttt{DEVICE\_TYPE\_ID} denotes whether this events happens on desktop, mobile or tablet. \texttt{LAYOUT\_ID} denotes a specific ads size and \texttt{AD\_POSITION\_ID} denotes the position for ads in web page.

Furthermore, for both data sets we filter out all features which appear less than $\tau$ times in the training set and replace them by a \texttt{NULL} feature, where $\tau$ is set to 20 for Criteo data set and 10 for Oath data set. The statistics of the two data sets are shown in Table~\ref{table:datasets}.

\begin{table}
\centering
    \begin{tabular}{ | c | c | r | c | c |} 
    \hline
    \multicolumn{2}{|c|}{Data set} & Samples & Fields &  Features \\
    \hline
    \multirow{3}{*}{Criteo} & Train & 27,502,713 & 26 & 399,784 \\
    \cline{2-5}
    & Validation & 9,168,820 & 26 & 399,654 \\
    \cline{2-5}
    & Test & 9,169,084 & 26 & 399,688 \\
    \hline
    \multirow{3}{*}{Oath} & Train & 24,885,731 & 15 & 156,401 \\
    \cline{2-5}
    & Validation & 7,990,874 & 15 & 101,217 \\
    \cline{2-5}
    & Test & 8,635,361 & 15  & 100,515 \\
    \hline
 	\end{tabular}
    \caption{Statistics of training, validation and test sets of Criteo and Oath data sets respectively.}
    \label{table:datasets}
\end{table}

\subsection{Implementations}
\label{subsec:implementation}

\begin{figure}[h!]
\centering
    \includegraphics[scale=0.34]{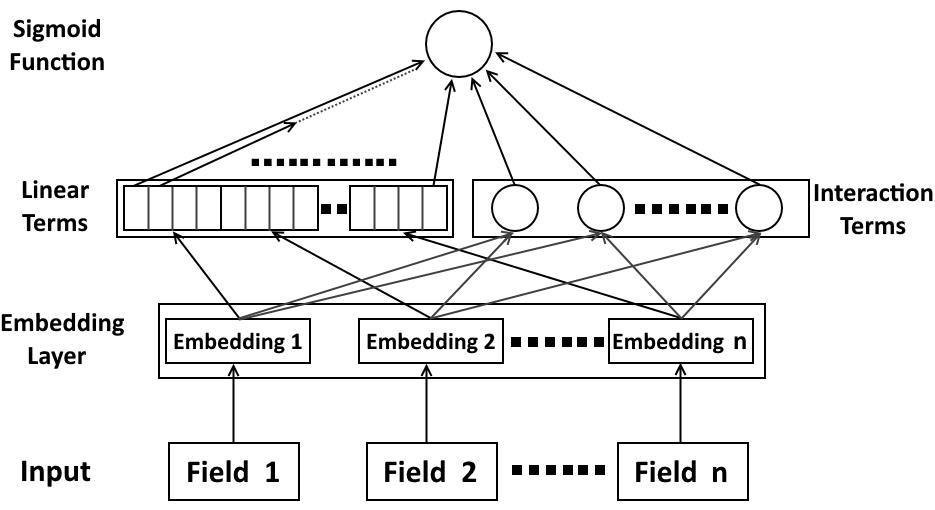}
    \caption{Implementation for FwFMs in Tensorflow.}
    \label{fig:tensorflow}
\end{figure}

We use LibLinear~\cite{fan2008liblinear} to train Poly2 with hashing tricks~\cite{weinberger2009feature} to hash feature conjunctions to a hashing space of $10^7$. All the other models are implemented in Tensorflow. We follow the implementation of LR and FMs in~\cite{qu2016product}, and implement FFMs by ourselves. The architecture of FwFMs in Tensorflow is shown in Figure~\ref{fig:tensorflow}.

The input is a sparse binary vector $\bm{x}_i \in \mathbb{R}^m$ with only $n$ non-zero entries since there are $n$ fields and each field has one and only one active feature for each sample. In the embedding layer, the input vector $\bm{x}_i$ is projected into $n$ embedding vectors. In the next layer the linear terms and interaction terms are computed from these $n$ latent vectors. The linear terms layer simply concatenates the latent vectors of all active features. The interaction terms layer calculates the dot product $\langle \bm{v}_i, \bm{v}_j \rangle$ between embedding vectors of all pairs of active features. Then every node in the linear terms layer and interaction terms layer will be connected to the final output node, which will sum up the inputs from both linear terms layer and interaction terms layer with weights. Note that in FwFMs\_FeLV, the weights between the linear terms layer and the final output is $w_{i}$ while for FwFMs\_FeLV and FwFM\_FiLV the weights are $\bm{w}_i$ and $\bm{w}_{F(i)}$ respectively. The weights between the interaction terms layer and the output node are $r_{F(i), F(j)}$.

\begin{table*}
\centering
\begin{subtable}{1\textwidth}
\sisetup{table-format=4.0} 
\centering
\begin{tabular}{| c | c | c | c | c |} 
\hline
	\multirow{2}{*}{Models} & \multirow{2}{*}{Parameters} & \multicolumn{3}{c|}{AUC} \\ \cline{3-5}
    & & Training & Validation & Test \\
	\hline
	LR & $\eta = 1e-4, \lambda = 1e-7, t = 15$ & 0.8595 & 0.8503 & 0.8503 \\
	\hline
	Poly2 & $s=7, c=2$ & 0.8652 & 0.8542 & 0.8523  \\
    \hline
	FMs & $\eta = 5e-4, \lambda = 1e-6, k = 10, t = 10 $ & 0.8768 & 0.8628 & 0.8583  \\
	\hline
	FFMs & $\eta = 1e-4, \lambda = 1e-7, k = 10, t = 3$ & \textbf{0.8833} & \textbf{0.8660} & \textbf{0.8624} \\
	\hline
	FwFMs & $\eta = 1e-4, \lambda = 1e-5, k = 10, t = 15$ & 0.8827 & 0.8659 & 0.8614 \\
    \hline
    \end{tabular}
	\caption{Oath data set}
\end{subtable}

\bigskip

\begin{subtable}{1\textwidth}
\sisetup{table-format=-1.2}   
\centering
	\begin{tabular}{| c | c | c | c | c |} 
    \hline
	\multirow{2}{*}{Models} & \multirow{2}{*}{Parameters} & \multicolumn{3}{c|}{AUC} \\ \cline{3-5}
    & & Training & Validation & Test \\
	\hline
	LR & $\eta = 5e-5 , \lambda = 1e-6, t = 14 $ & 0.7716 & 0.7657 & 0.7654 \\
	\hline
	Poly2 & $s=7, c=2 $ & 0.7847 & 0.7718 & 0.7710 \\
    \hline
	FMs & $\eta = 1e-4, \lambda = 1e-6, k = 10, t = 10 $ & 0.7925 & 0.7759 & 0.7761\\
	\hline
	FFMs & $\eta = 5e-4, \lambda = 1e-7, k = 10, t = 3 $ & \textbf{0.7989} & \textbf{0.7781} & \textbf{0.7768} \\
	\hline
	FwFMs & $\eta = 1e-4, \lambda = 1e-6, k = 10, t = 8$ & 0.7941 & 0.7772 & 0.7764 \\
    \hline
    \end{tabular}
	\caption{Criteo data set}
\end{subtable}
\caption{Comparison among models on Criteo and Oath CTR data sets.}
\label{table:fwfm-vs-others}
\end{table*}

\subsection{Performance Comparisons}

\subsubsection{Comparison of FwFMs with Existing Models}
\label{subsec:performance_comparison}

In this section we will conduct performance evaluation for FwFMs with original linear terms, i.e., FwFMs\_LW. We will compare it with LR, Poly2, FMs and FFMs on two data sets mentioned above. For the parameters such as regularization coefficient $\lambda$, and learning rate $\eta$ in all models, we select those which leads to the best performance on the validation set and then use them in the evaluation on test set. Experiment results can be found in Table~\ref{table:fwfm-vs-others}.

We observe that FwFMs can achieve better performance than LR, Poly2 and FMs on both data sets. The improvement comes from the fact that FwFMs explicitly model the different interaction strengths of field pairs, which would be discussed in Section~\ref{sec:analysis} in more details. FFMs can always get the best performance on training and validation sets. However, FwFMs' performance on the both test sets are quite competitive with that of FwFMs. It suggests that FFMs are more vulnerable to overfittings than FwFMs.

\subsubsection{Comparison of FwFMs and FFMs using the same number of parameters} 
\label{subsec:same_parameter_number}

One critical drawback of using FFMs is that their number of parameters is in the order of $O(mnK)$, which would be too large to fit in memory. There are two solutions to reduce the number of parameters in FFMs~\cite{juan2017field}: use a smaller $K$ or use hashing tricks with a small hashing space $H$. Juan et. al.~\cite{juan2017field} proposed to use $K_{FFMs}=2$  for FFMs to get a good trade-off between prediction accuracy and the number of parameters. This reduce the number of parameters to $(n-1)m K_{FFMs}$. Juan et. al.~\cite{juan2017field} also proposed to further reduce the number of parameters of FFMs by using hashing tricks with a small hashing space $H_{FFMs}$ to reduce it to $(n-1) m_{FFMs} K_{FFMs}$. In this section, we make a fair comparison of FwFMs with FFMs using the same number of parameters by choosing proper $k_{FFMs}$ and $H_{FFMs}$ so that the number of parameters of FFMs and FwFMs are the same, i.e., $(n-1) H_{FFMs} K_{FFMs} = m K_{FwFMs}$, we do not count the parameters from the linear terms since they are negligible compared with the number of interaction terms. We choose $K_{FwFMs}=10$ and $K_{FFMs}=2$ and $K_{FFMs}=4$ as described in~\cite{juan2017field}. The experimental results are shown in Table~\ref{table:fwFm-vw-ffm}. 

We observe that when using the same number of parameters, FwFMs get better performance on test sets of both Criteo and Oath data sets, with a lift of 0.70\% and 0.45\% respectively. We conclude that FFMs make lots of compromise on the prediction performance when reducing the feature numbers therefore FwFMs can outperform them significantly in such cases. 

\begin{table*}
\centering
    \begin{tabular}{| c | c | c | c | c | c | c |} 
    \hline
      \multirow{2}{*}{Model} & \multicolumn{3}{c|}{Oath data set} & \multicolumn{3}{c|}{Criteo data set} \\
      \cline{2-7}
      & Training & Validation & Test & Training & Validation & Test  \\
      \hline
      FFMs($K=2,H=\frac{10}{14 \cdot 2}m$) & 0.8743 & 0.8589 & 0.8543 & 0.7817 & 0.7716 & 0.7719\\
      \hline
      FFMs($K=4,H=\frac{10}{14 \cdot 4}m$) & 0.8708 & 0.8528 & 0.8418 & 0.7697 & 0.7643 & 0.7641\\
      \hline
      FwFMs & \textbf{0.8827} & \textbf{0.8659} & \textbf{0.8614} & \textbf{0.7941} & \textbf{0.7772} & \textbf{0.7764} \\
      \hline
    \end{tabular}
\caption{Comparison of FFMs with FwFMs using same number of parameters. $K$ is embedding vector dimension and $H$ is the hashing space for hashing tricks.}
\label{table:fwFm-vw-ffm}
\end{table*}

\subsubsection{FwFMs with Different Linear Terms}
\label{subsec:performance_linear_terms}

We conduct experiments to compare FwFMs with three different kinds of linear terms as mentioned in Section~\ref{subsec:linear-terms}. Table~\ref{table:linear-terms} lists the performance comparison between them. We observe that, FwFMs\_LW and FwFMs\_FeLV can achieve better performance on the training and validation set than FwFMs\_FiLV. The reason is that those two models have more number of parameters in the linear weights($m$ and $mK$) than FwFMs\_FiLV($nK$), so they can fit the training and validation set better than FwFMs\_FiLV. However, FwFMs\_FiLV get the best results on the test set, which suggests that it has better generalization performance. Furthermore, when we compare FwFMs\_FiLV with FFMs using the same number of parameters, the AUC lift on Oath data set and Criteo data set are 0.92\% and 0.47\% respectively.

\begin{table*}
\centering
	\begin{tabular}{| c | c | c | c | c | c | c |}
    \hline
    \multirow{2}{*}{Model} & \multicolumn{3}{c|}{Oath data set} & \multicolumn{3}{c|}{Criteo data set} \\
      \cline{2-7}
      & Training & Validation & Test & Training & Validation & Test  \\
      \hline 
    FwFMs\_LW & \textbf{0.8827} & 0.8659 & 0.8614 & 0.7941 & 0.7772 & 0.7764 \\
    \hline
    FwFMs\_FeLV & 0.8829 & \textbf{0.8665} & 0.8623 & \textbf{0.7945} & \textbf{0.7774} & 0.7763 \\
    \hline
    FwFMs\_FiLV & 0.8799 & 0.8643 & \textbf{0.8635} & 0.7917 & 0.7766 & \textbf{0.7766} \\
    \hline
	\end{tabular}
    \caption{Performance of FwFMs with different linear terms.}
    \label{table:linear-terms}
\end{table*}

\subsection{Hyper-parameter Tuning}
\label{subsec:parameter_tuning}

In this section we will show the impact of regularization coefficient $\lambda$, embedding vector dimension $K$ and learning rate $\eta$ on FwFMs. All following evaluations are done for FwFMs\_FiLV model on Oath validation set.

\subsubsection{Regularization}

\begin{figure}
\includegraphics[width=0.9\columnwidth]{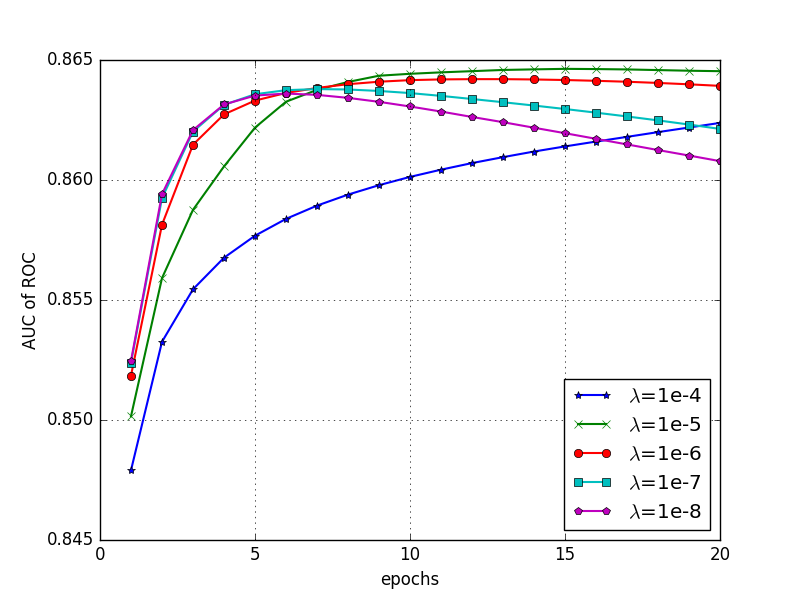}
    \caption{Impact of regularization coefficient $\lambda$ on FwFMs.} 
    \label{fig:regularization}
\end{figure}

We add $L_2$ regularizations of all parameters in FwFMs to the loss function to prevent over-fitting. Figure~\ref{fig:regularization} shows the AUC on validation set using different $\lambda$. We get the best performance on validation set using $\lambda=1e-5$.

\subsubsection{Learning Rate and Embedding Vector Dimension} 

\begin{figure}
    \includegraphics[width=0.9\columnwidth]{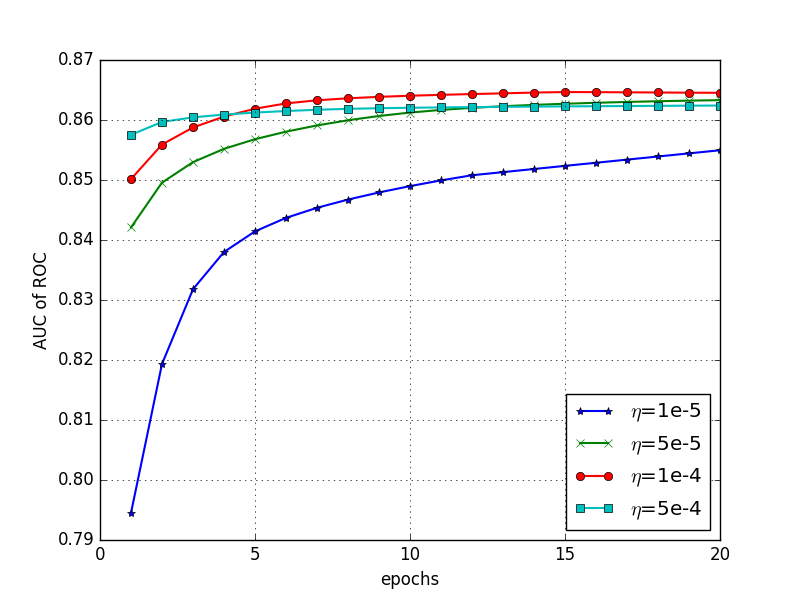}
    \caption{Impact of learning rate $\eta$ on FwFMs.}
    \label{fig:learning-rate}
\end{figure}

We have done experiments to check the impact of learning rate $\eta$ to the performance of FwFMs, and the results are shown in Figure~\ref{fig:learning-rate}. It shows that by using a small $\eta$, we can keep improving the performance on validation set slowly in the first 20 epochs, while using a large $\eta$ will improve the performance quickly and then lead to over-fitting. In all experiments on Oath data set we choose $\eta = 1e-4$.

We conduct experiments to investigate the impact of embedding vector dimension $K$ on the performance of FwFMs. Table~\ref{table:embedding-size} shows that the AUC changes only a little when we use different $K$. We choose $K=10$ in FwFMs since it gets best trade-off between performance and training time.

\begin{table}
\centering
	\begin{tabular}{| c | c | c | c |}
    \hline
    $k$ & Train AUC & Validation AUC & Training time (s) \\
    \hline
    5 & 0.8794 & 0.8621 &  320 \\
    \hline
	10 & 0.8827 & 0.8659 & 497\\
    \hline
	15 & 0.8820 & 0.8644 & 544 \\
    \hline
	20 & 0.8822 & 0.8640 & 636 \\
    \hline
	30 & 0.8818 & 0.8647 & 848 \\
    \hline
	50 & 0.8830 & 0.8652 & 1113 \\
    \hline
	100 & 0.8830 & 0.8646 & 1728 \\
    \hline
	200 & 0.8825 & 0.8646 & 3250\\
    \hline
	\end{tabular}
    \caption{Comparison of different embedding vector dimensions $K$}
    \label{table:embedding-size}
\end{table}

\section{Study of Learned Field Interaction Strengths}\label{sec:analysis}
In this section, we compare FMs, FFMs and FwFMs in terms of their ability to capture interaction strengths of different field pairs. Our result shows that FwFMs can model the interaction strength much better than FMs and FFMs do thanks to the field pair interaction weight $r_{F_k, F_l}$. 

\begin{figure*}[ht]
  \captionsetup{width=.8\linewidth}
  \centering
  \begin{subfigure}[t]{\columnwidth}
    \centering
    \captionsetup{width=.8\linewidth}
    \includegraphics[width=\columnwidth]{images/Heat_Map_MI.png} 
    \caption{Heat map of mutual informations between field pairs and labels.}
    \label{subfig:heat-map-mi}
    \vspace{4ex}
  \end{subfigure}
  \begin{subfigure}[t]{\columnwidth}
    \centering
    \captionsetup{width=.8\linewidth}
    \includegraphics[width=\columnwidth]{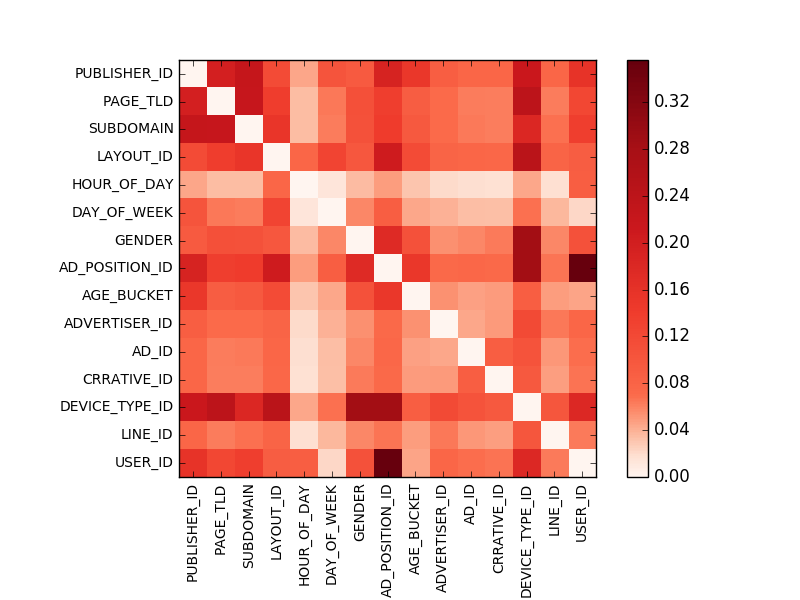} 
    \caption{Heat map of learned field interaction strengths of FMs. Pearson correlation coefficient with mutual informations: -0.1210.}
    \label{subfig:heat-map-fm}
    \vspace{4ex}
  \end{subfigure} 
  \begin{subfigure}[t]{\columnwidth}
    \centering
    \captionsetup{width=.8\linewidth}
    \includegraphics[width=\columnwidth]{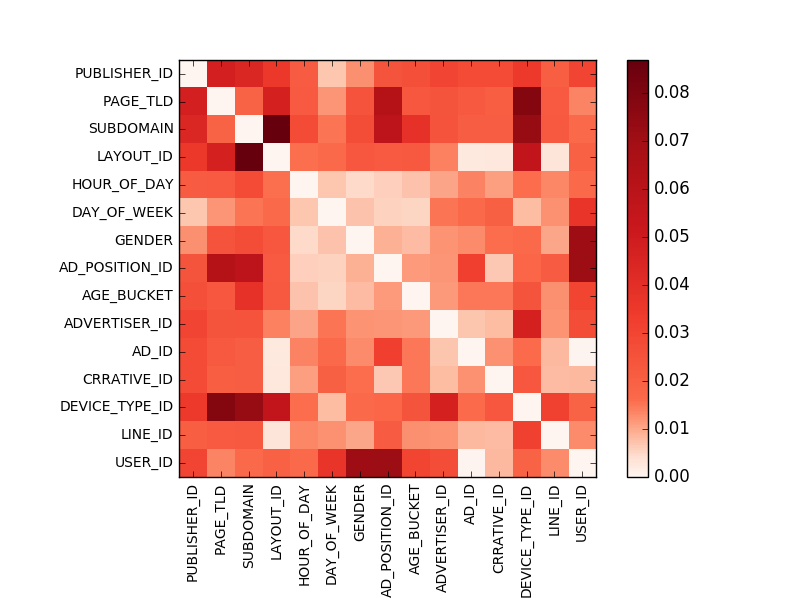} 
    \caption{Heat map of learned field interaction strengths of FFMs. Pearson correlation coefficient with mutual informations: 0.1544.}
    \label{subfig:heat-map-ffm}
    \vspace{4ex}
  \end{subfigure}
  \begin{subfigure}[t]{\columnwidth}
    \centering
    \captionsetup{width=.8\linewidth}
    \includegraphics[width=\columnwidth]{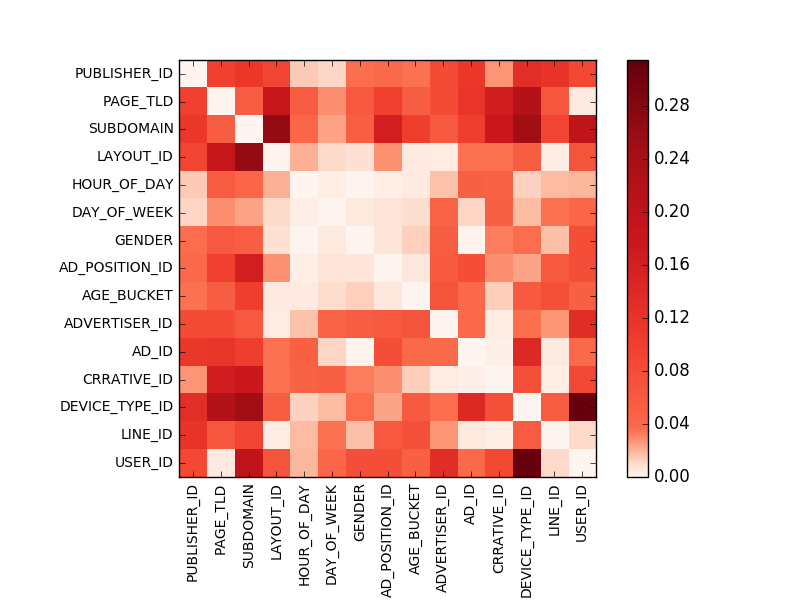} 
    \caption{Heat map of learned field interaction strengths of FwFMs. Pearson correlation coefficient with mutual informations: 0.4271.}
    \label{subfig:heat-map-fwfm}
    \vspace{4ex}
  \end{subfigure} 

  \caption{Heat maps of mutual informations (a) and learned field pair interaction strengths of FMs (b), FFMs (c) and FwFMs (d) on Oath CTR data set.}
  \label{fig:heat-map-all}
\end{figure*}


In Section~\ref{sec:interaction} the interaction strengths of field pairs are quantified by mutual information and visualized in heat map (Figure~\ref{subfig:heat-map-mi}). To measure the learned field interaction strengths, we define the following metric:
\begin{equation}\label{eq:interaction_metric}
	\frac{\sum_{(i,j) \in (F_k, F_l)} I(i,j) \cdot \#(i,j)}{\sum_{(i,j) \in (F_k, F_l)} \#(i,j)}
\end{equation}
where $\#(i,j)$ is the number of times feature pair $(i,j)$ appears in the training data, $I(i,j)$ is the learned strength of interaction between feature $i$ and $j$. For FMs $I(i,j) =  |\langle \bm{v}_{i}, \bm{v}_{j} \rangle|$ , for FFMs $I(i,j) = |\langle \bm{v}_{i, F_l}, \bm{v}_{j, F_k} \rangle|$, for FwFMs  $I(i,j) = |\langle \bm{v}_i, \bm{v}_j \rangle \cdot r_{F_k, F_l}|$. Note that we sum up the absolute values of the inner product terms otherwise positive values and negative values would counteract with each others.

If a model can capture the interaction strengths of different field pairs well, we expect its learned field interaction strengths to be close to the mutual 
information as shown in equation~\eqref{eq:MI}. To facilitate the comparison, in Figure~\ref{fig:heat-map-all} we plot the field interaction strengths learned by FMs, FFMs and FwFMs in the form of heatmap, along with heatmap of the mutual information between field pairs and labels. We also computed Pearson correlation coefficient to quantitatively measure how well the learned field interaction strengths match the mutual information.

From Figure~\ref{subfig:heat-map-fm} we observe that FMs learned interaction strengths quite different from mutual information. Some field pairs which have high mutual information with labels only get low interaction strengths, for example \texttt{(AD\_ID, SUBDOMAIN)}, \texttt{(CREATIVE\_ID, PAGE\_TLD)}. While some field pairs with low mutual information have high interaction strengths, such as \texttt{(LAYOUT\_ID, GENDER)}. The Pearson correlation coefficient between the learned field pair interaction strengths and mutual information is -0.1210. This is not surprising since FMs is not considering field information when modeling feature interaction strengths.

As shown in Figure~\ref{subfig:heat-map-ffm}, FFMs are able to learn interaction strengths more similar to mutual information. This is confirmed by a higher Pearson correlation coefficient of 0.1544, which is much better than FMs. 

For FwFMs, Figure~\ref{subfig:heat-map-fwfm} shows that the heat map of the learned interaction strengths are very similar to that of mutual information. For example, the field pairs that have highest mutual informations with labels, such as (\texttt{AD\_ID},\texttt{SUBDOMAIN}),
(\texttt{AD\_ID},\texttt{PAGE\_TLD}), and
(\texttt{CREATIVE\_ID},\texttt{PAGE\_TLD}), also have higher interaction strengths.
For fields such as \texttt{LAYOUT\_ID}, \texttt{HOUR\_OF\_DAY}, \texttt{DAY\_OF\_WEEK}, \texttt{GENDER} and \texttt{AD\_POSITION\_ID}, field pairs between any of them have low mutual information with labels as well as low interaction strengths. The Pearson correlation coefficient with mutual information is 0.4271, which is much better than FMs and FFMs.

\begin{figure*}[ht]
	\captionsetup{width=.8\linewidth}
	\begin{subfigure}[t]{\columnwidth}
    \centering
    \captionsetup{width=.8\linewidth}
    \includegraphics[width=\columnwidth]{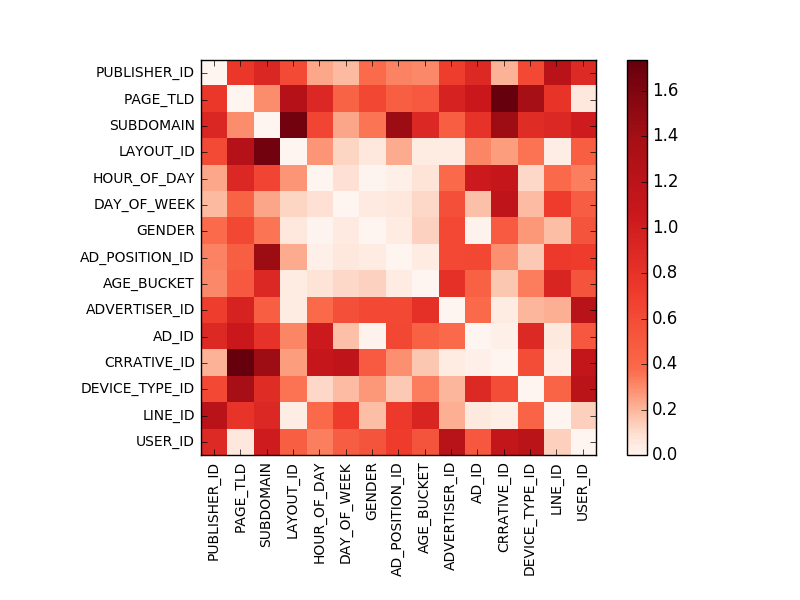} 
    \caption{Heat map of field interaction weight $r_{F_k, F_l}$. Its Pearson correlation coefficient with mutual information is 0.5554.}
    \label{subfig:heat-map-r}
    \vspace{4ex}
  \end{subfigure}
  \begin{subfigure}[t]{\columnwidth}
    \centering
    \captionsetup{width=.8\linewidth}
    \includegraphics[width=\columnwidth]{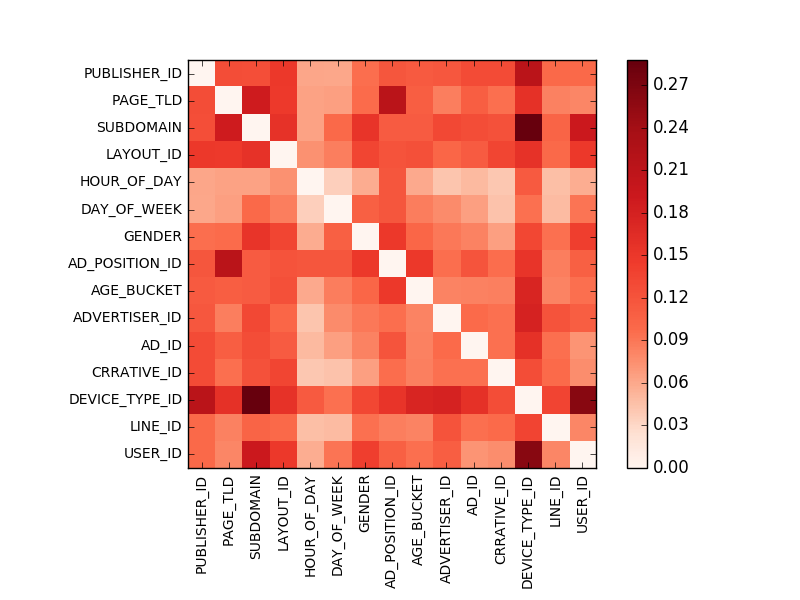} 
    \caption{Heat map of learned field interaction strengths of FwFMs without $r_{F_k, F_l}$. Its Pearson correlation coefficient with mutual information is 0.0522.}
    \label{subfig:heat-map-fwfm-without-r}
    \vspace{4ex}
  \end{subfigure} 
  \caption{Heat map of learned field interaction weight $r_{F_k, F_l}$ , and heat map of learned interaction strengths between field pairs of FwFMs without $r_{F_k, F_l}$}
  \label{fig:heat-map-fwfms}
\end{figure*}

To understand the importance of the weights $r_{F_k, F_l}$ in modeling the interaction strength of field pairs, we plot the heap map of $r_{F_k, F_l}$, as well as learned field interaction strengths of FwFMs without $r_{F_k, F_l}$ in Figure~\ref{fig:heat-map-fwfms}.

The heat map of $r_{F_k, F_l}$ (Figure \ref{subfig:heat-map-r}) is very similar to that of mutual information (Figure \ref{subfig:heat-map-mi}) and learned field interaction strengths of FwFMs (Figure \ref{subfig:heat-map-fwfm}). The corresponding Pearson correlation coefficient with mutual information is 0.5554. This implies that $r_{F_k, F_l}$ can indeed learn different interaction strengths of different field pairs. On the other hand, the learned field interaction strengths without $r_{F_k, F_l}$ correlates poorly with mutual information, as depicted in Figure~\ref{subfig:heat-map-fwfm-without-r}. Its Pearson correlation coefficient with mutual information is only 0.0522.

\section{Related Work}\label{sec:related}
CTR prediction is an essential task in online display advertising. Many models have been proposed to resolve this problem such as Logistic Regression (LR) ~\cite{richardson2007predicting,chapelle2015simple,mcmahan2013ad}, Polynomial-2 (Poly2)  ~\cite{chang2010training}, tree-based models~\cite{he2014practical}, tensor-based models~\cite{rendle2010pairwise}, Bayesian models~\cite{graepel2010web}, and Field-aware Factorization Machines (FFMs)~\cite{juan2016field,juan2017field}. Recently, deep learning for CTR prediction also attracted much research attention~\cite{cheng2016wide,zhang2016deep,qu2016product,guo2017deepfm,shan2016deep,he2017neural,wang2017deep}. However, only FFMs explicitly model different feature interactions from different fields in the multi-field categorical data, which we have shown to be very effective for CTR prediction.

FFMs extended the ideas of Factorization Machines (FMs)~\cite{rendle2010factorization}. The latter have been proved to be very successful in recommender systems. Recommendation is a different but very closely related topic with CTR prediction. Collaborative filtering (CF) techniques such as matrix factorization~\cite{koren2009matrix} have been established as standard approaches for various recommendation tasks. However, FMs have been shown to outperform matrix factorization in many recommendation tasks~\cite{juan2017field,thai2012using} and they are also capable of effectively resolving cold-start problems~\cite{aharon2013off}.



\section{Conclusion}\label{sec:conclusion}
In this paper, we propose Field-weighted Factorization Machines (FwFMs) for CTR prediction in online display advertising. We show that FwFMs are competitive to FFMs with significantly less parameters. When using the same number of parameters, FwFMs can achieve consistently better performance than FFMs. We also introduce a novel linear term representation to augment FwFMs so that their performance can be further improved. Finally, comprehensive analysis on real-world data sets also verifies that FwFMs can indeed learn different feature interaction strengths from different field pairs. There are many potential directions for future research. For example, it is interesting to see whether there are other alternatives to model different field interactions rather than learning one weight for each field pair. Another direction is to explore the possibility of combining deep learning with FwFMs since the combination of deep neural networks and FMs has shown promising results in CTR prediction \cite{zhang2016deep,guo2017deepfm,he2017neural,qu2016product}.

\bibliographystyle{ACM-Reference-Format}
\bibliography{ms}


\begin{thebibliography}{26}


\ifx \showCODEN    \undefined \def \showCODEN     #1{\unskip}     \fi
\ifx \showDOI      \undefined \def \showDOI       #1{#1}\fi
\ifx \showISBNx    \undefined \def \showISBNx     #1{\unskip}     \fi
\ifx \showISBNxiii \undefined \def \showISBNxiii  #1{\unskip}     \fi
\ifx \showISSN     \undefined \def \showISSN      #1{\unskip}     \fi
\ifx \showLCCN     \undefined \def \showLCCN      #1{\unskip}     \fi
\ifx \shownote     \undefined \def \shownote      #1{#1}          \fi
\ifx \showarticletitle \undefined \def \showarticletitle #1{#1}   \fi
\ifx \showURL      \undefined \def \showURL       {\relax}        \fi
\providecommand\bibfield[2]{#2}
\providecommand\bibinfo[2]{#2}
\providecommand\natexlab[1]{#1}
\providecommand\showeprint[2][]{arXiv:#2}

\bibitem[\protect\citeauthoryear{Aharon, Aizenberg, Bortnikov, Lempel, Adadi,
  Benyamini, Levin, Roth, and Serfaty}{Aharon et~al\mbox{.}}{2013}]%
        {aharon2013off}
\bibfield{author}{\bibinfo{person}{Michal Aharon}, \bibinfo{person}{Natalie
  Aizenberg}, \bibinfo{person}{Edward Bortnikov}, \bibinfo{person}{Ronny
  Lempel}, \bibinfo{person}{Roi Adadi}, \bibinfo{person}{Tomer Benyamini},
  \bibinfo{person}{Liron Levin}, \bibinfo{person}{Ran Roth}, {and}
  \bibinfo{person}{Ohad Serfaty}.} \bibinfo{year}{2013}\natexlab{}.
\newblock \showarticletitle{OFF-set: one-pass factorization of feature sets for
  online recommendation in persistent cold start settings}. In
  \bibinfo{booktitle}{\emph{Proceedings of the 7th ACM Conference on
  Recommender Systems}}. ACM, \bibinfo{pages}{375--378}.
\newblock


\bibitem[\protect\citeauthoryear{Bureau}{Bureau}{2016}]%
        {iab-report}
\bibfield{author}{\bibinfo{person}{Interactive~Advertising Bureau}.}
  \bibinfo{year}{2016}\natexlab{}.
\newblock \bibinfo{title}{IAB internet advertising revenue report}.
\newblock   (\bibinfo{year}{2016}).
\newblock
\urldef\tempurl%
\url{https://www.iab.com/wp-content/uploads/2016/04/IAB_Internet_Advertising_Revenue_Report_FY_2016.pdf}
\showURL{%
\tempurl}


\bibitem[\protect\citeauthoryear{Chang, Hsieh, Chang, Ringgaard, and Lin}{Chang
  et~al\mbox{.}}{2010}]%
        {chang2010training}
\bibfield{author}{\bibinfo{person}{Yin-Wen Chang}, \bibinfo{person}{Cho-Jui
  Hsieh}, \bibinfo{person}{Kai-Wei Chang}, \bibinfo{person}{Michael Ringgaard},
  {and} \bibinfo{person}{Chih-Jen Lin}.} \bibinfo{year}{2010}\natexlab{}.
\newblock \showarticletitle{Training and testing low-degree polynomial data
  mappings via linear SVM}.
\newblock \bibinfo{journal}{\emph{Journal of Machine Learning Research}}
  \bibinfo{volume}{11}, \bibinfo{number}{Apr} (\bibinfo{year}{2010}),
  \bibinfo{pages}{1471--1490}.
\newblock


\bibitem[\protect\citeauthoryear{Chapelle, Manavoglu, and Rosales}{Chapelle
  et~al\mbox{.}}{2015}]%
        {chapelle2015simple}
\bibfield{author}{\bibinfo{person}{Olivier Chapelle}, \bibinfo{person}{Eren
  Manavoglu}, {and} \bibinfo{person}{Romer Rosales}.}
  \bibinfo{year}{2015}\natexlab{}.
\newblock \showarticletitle{Simple and scalable response prediction for display
  advertising}.
\newblock \bibinfo{journal}{\emph{ACM Transactions on Intelligent Systems and
  Technology (TIST)}} \bibinfo{volume}{5}, \bibinfo{number}{4}
  (\bibinfo{year}{2015}), \bibinfo{pages}{61}.
\newblock


\bibitem[\protect\citeauthoryear{Cheng, Koc, Harmsen, Shaked, Chandra, Aradhye,
  Anderson, Corrado, Chai, Ispir, et~al\mbox{.}}{Cheng et~al\mbox{.}}{2016}]%
        {cheng2016wide}
\bibfield{author}{\bibinfo{person}{Heng-Tze Cheng}, \bibinfo{person}{Levent
  Koc}, \bibinfo{person}{Jeremiah Harmsen}, \bibinfo{person}{Tal Shaked},
  \bibinfo{person}{Tushar Chandra}, \bibinfo{person}{Hrishi Aradhye},
  \bibinfo{person}{Glen Anderson}, \bibinfo{person}{Greg Corrado},
  \bibinfo{person}{Wei Chai}, \bibinfo{person}{Mustafa Ispir}, {et~al\mbox{.}}}
  \bibinfo{year}{2016}\natexlab{}.
\newblock \showarticletitle{Wide \& deep learning for recommender systems}. In
  \bibinfo{booktitle}{\emph{Proceedings of the 1st Workshop on Deep Learning
  for Recommender Systems}}. ACM, \bibinfo{pages}{7--10}.
\newblock


\bibitem[\protect\citeauthoryear{Cover and Thomas}{Cover and Thomas}{2012}]%
        {cover2012elements}
\bibfield{author}{\bibinfo{person}{Thomas~M Cover} {and} \bibinfo{person}{Joy~A
  Thomas}.} \bibinfo{year}{2012}\natexlab{}.
\newblock \bibinfo{booktitle}{\emph{Elements of information theory}}.
\newblock \bibinfo{publisher}{John Wiley \& Sons}.
\newblock


\bibitem[\protect\citeauthoryear{Fan, Chang, Hsieh, Wang, and Lin}{Fan
  et~al\mbox{.}}{2008}]%
        {fan2008liblinear}
\bibfield{author}{\bibinfo{person}{Rong-En Fan}, \bibinfo{person}{Kai-Wei
  Chang}, \bibinfo{person}{Cho-Jui Hsieh}, \bibinfo{person}{Xiang-Rui Wang},
  {and} \bibinfo{person}{Chih-Jen Lin}.} \bibinfo{year}{2008}\natexlab{}.
\newblock \showarticletitle{LIBLINEAR: A library for large linear
  classification}.
\newblock \bibinfo{journal}{\emph{Journal of machine learning research}}
  \bibinfo{volume}{9}, \bibinfo{number}{Aug} (\bibinfo{year}{2008}),
  \bibinfo{pages}{1871--1874}.
\newblock


\bibitem[\protect\citeauthoryear{Graepel, Candela, Borchert, and
  Herbrich}{Graepel et~al\mbox{.}}{2010}]%
        {graepel2010web}
\bibfield{author}{\bibinfo{person}{Thore Graepel}, \bibinfo{person}{Joaquin~Q
  Candela}, \bibinfo{person}{Thomas Borchert}, {and} \bibinfo{person}{Ralf
  Herbrich}.} \bibinfo{year}{2010}\natexlab{}.
\newblock \showarticletitle{Web-scale bayesian click-through rate prediction
  for sponsored search advertising in microsoft's bing search engine}. In
  \bibinfo{booktitle}{\emph{Proceedings of the 27th international conference on
  machine learning (ICML-10)}}. \bibinfo{pages}{13--20}.
\newblock


\bibitem[\protect\citeauthoryear{Guo, Tang, Ye, Li, and He}{Guo
  et~al\mbox{.}}{2017}]%
        {guo2017deepfm}
\bibfield{author}{\bibinfo{person}{Huifeng Guo}, \bibinfo{person}{Ruiming
  Tang}, \bibinfo{person}{Yunming Ye}, \bibinfo{person}{Zhenguo Li}, {and}
  \bibinfo{person}{Xiuqiang He}.} \bibinfo{year}{2017}\natexlab{}.
\newblock \showarticletitle{DeepFM: A Factorization-Machine based Neural
  Network for CTR Prediction}.
\newblock \bibinfo{journal}{\emph{arXiv preprint arXiv:1703.04247}}
  (\bibinfo{year}{2017}).
\newblock


\bibitem[\protect\citeauthoryear{He and Chua}{He and Chua}{2017}]%
        {he2017neural}
\bibfield{author}{\bibinfo{person}{Xiangnan He} {and} \bibinfo{person}{Tat-Seng
  Chua}.} \bibinfo{year}{2017}\natexlab{}.
\newblock \showarticletitle{Neural Factorization Machines for Sparse Predictive
  Analytics}.
\newblock  (\bibinfo{year}{2017}).
\newblock


\bibitem[\protect\citeauthoryear{He, Pan, Jin, Xu, Liu, Xu, Shi, Atallah,
  Herbrich, Bowers, et~al\mbox{.}}{He et~al\mbox{.}}{2014}]%
        {he2014practical}
\bibfield{author}{\bibinfo{person}{Xinran He}, \bibinfo{person}{Junfeng Pan},
  \bibinfo{person}{Ou Jin}, \bibinfo{person}{Tianbing Xu}, \bibinfo{person}{Bo
  Liu}, \bibinfo{person}{Tao Xu}, \bibinfo{person}{Yanxin Shi},
  \bibinfo{person}{Antoine Atallah}, \bibinfo{person}{Ralf Herbrich},
  \bibinfo{person}{Stuart Bowers}, {et~al\mbox{.}}}
  \bibinfo{year}{2014}\natexlab{}.
\newblock \showarticletitle{Practical lessons from predicting clicks on ads at
  facebook}. In \bibinfo{booktitle}{\emph{Proceedings of the Eighth
  International Workshop on Data Mining for Online Advertising}}. ACM,
  \bibinfo{pages}{1--9}.
\newblock


\bibitem[\protect\citeauthoryear{Juan, Lefortier, and Chapelle}{Juan
  et~al\mbox{.}}{2017}]%
        {juan2017field}
\bibfield{author}{\bibinfo{person}{Yuchin Juan}, \bibinfo{person}{Damien
  Lefortier}, {and} \bibinfo{person}{Olivier Chapelle}.}
  \bibinfo{year}{2017}\natexlab{}.
\newblock \showarticletitle{Field-aware factorization machines in a real-world
  online advertising system}. In \bibinfo{booktitle}{\emph{Proceedings of the
  26th International Conference on World Wide Web Companion}}. International
  World Wide Web Conferences Steering Committee, \bibinfo{pages}{680--688}.
\newblock


\bibitem[\protect\citeauthoryear{Juan, Zhuang, Chin, and Lin}{Juan
  et~al\mbox{.}}{2016}]%
        {juan2016field}
\bibfield{author}{\bibinfo{person}{Yuchin Juan}, \bibinfo{person}{Yong Zhuang},
  \bibinfo{person}{Wei-Sheng Chin}, {and} \bibinfo{person}{Chih-Jen Lin}.}
  \bibinfo{year}{2016}\natexlab{}.
\newblock \showarticletitle{Field-aware factorization machines for CTR
  prediction}. In \bibinfo{booktitle}{\emph{Proceedings of the 10th ACM
  Conference on Recommender Systems}}. ACM, \bibinfo{pages}{43--50}.
\newblock


\bibitem[\protect\citeauthoryear{Koren, Bell, and Volinsky}{Koren
  et~al\mbox{.}}{2009}]%
        {koren2009matrix}
\bibfield{author}{\bibinfo{person}{Yehuda Koren}, \bibinfo{person}{Robert
  Bell}, {and} \bibinfo{person}{Chris Volinsky}.}
  \bibinfo{year}{2009}\natexlab{}.
\newblock \showarticletitle{Matrix factorization techniques for recommender
  systems}.
\newblock \bibinfo{journal}{\emph{Computer}} \bibinfo{volume}{42},
  \bibinfo{number}{8} (\bibinfo{year}{2009}).
\newblock


\bibitem[\protect\citeauthoryear{Labs}{Labs}{2014}]%
        {criteo-display-ad-challenge}
\bibfield{author}{\bibinfo{person}{Criteo Labs}.}
  \bibinfo{year}{2014}\natexlab{}.
\newblock \bibinfo{title}{Display Advertising Challenge}.
\newblock   (\bibinfo{year}{2014}).
\newblock
\urldef\tempurl%
\url{https://www.kaggle.com/c/criteo-display-ad-challenge}
\showURL{%
\tempurl}


\bibitem[\protect\citeauthoryear{McMahan, Holt, Sculley, Young, Ebner, Grady,
  Nie, Phillips, Davydov, Golovin, et~al\mbox{.}}{McMahan
  et~al\mbox{.}}{2013}]%
        {mcmahan2013ad}
\bibfield{author}{\bibinfo{person}{H~Brendan McMahan}, \bibinfo{person}{Gary
  Holt}, \bibinfo{person}{David Sculley}, \bibinfo{person}{Michael Young},
  \bibinfo{person}{Dietmar Ebner}, \bibinfo{person}{Julian Grady},
  \bibinfo{person}{Lan Nie}, \bibinfo{person}{Todd Phillips},
  \bibinfo{person}{Eugene Davydov}, \bibinfo{person}{Daniel Golovin},
  {et~al\mbox{.}}} \bibinfo{year}{2013}\natexlab{}.
\newblock \showarticletitle{Ad click prediction: a view from the trenches}. In
  \bibinfo{booktitle}{\emph{Proceedings of the 19th ACM SIGKDD international
  conference on Knowledge discovery and data mining}}. ACM,
  \bibinfo{pages}{1222--1230}.
\newblock


\bibitem[\protect\citeauthoryear{Qu, Cai, Ren, Zhang, Yu, Wen, and Wang}{Qu
  et~al\mbox{.}}{2016}]%
        {qu2016product}
\bibfield{author}{\bibinfo{person}{Yanru Qu}, \bibinfo{person}{Han Cai},
  \bibinfo{person}{Kan Ren}, \bibinfo{person}{Weinan Zhang},
  \bibinfo{person}{Yong Yu}, \bibinfo{person}{Ying Wen}, {and}
  \bibinfo{person}{Jun Wang}.} \bibinfo{year}{2016}\natexlab{}.
\newblock \showarticletitle{Product-based neural networks for user response
  prediction}. In \bibinfo{booktitle}{\emph{Data Mining (ICDM), 2016 IEEE 16th
  International Conference on}}. IEEE, \bibinfo{pages}{1149--1154}.
\newblock


\bibitem[\protect\citeauthoryear{Rendle}{Rendle}{2010}]%
        {rendle2010factorization}
\bibfield{author}{\bibinfo{person}{Steffen Rendle}.}
  \bibinfo{year}{2010}\natexlab{}.
\newblock \showarticletitle{Factorization machines}. In
  \bibinfo{booktitle}{\emph{Data Mining (ICDM), 2010 IEEE 10th International
  Conference on}}. IEEE, \bibinfo{pages}{995--1000}.
\newblock


\bibitem[\protect\citeauthoryear{Rendle}{Rendle}{2012}]%
        {rendle2012factorization}
\bibfield{author}{\bibinfo{person}{Steffen Rendle}.}
  \bibinfo{year}{2012}\natexlab{}.
\newblock \showarticletitle{Factorization machines with libfm}.
\newblock \bibinfo{journal}{\emph{ACM Transactions on Intelligent Systems and
  Technology (TIST)}} \bibinfo{volume}{3}, \bibinfo{number}{3}
  (\bibinfo{year}{2012}), \bibinfo{pages}{57}.
\newblock


\bibitem[\protect\citeauthoryear{Rendle and Schmidt-Thieme}{Rendle and
  Schmidt-Thieme}{2010}]%
        {rendle2010pairwise}
\bibfield{author}{\bibinfo{person}{Steffen Rendle} {and} \bibinfo{person}{Lars
  Schmidt-Thieme}.} \bibinfo{year}{2010}\natexlab{}.
\newblock \showarticletitle{Pairwise interaction tensor factorization for
  personalized tag recommendation}. In \bibinfo{booktitle}{\emph{Proceedings of
  the third ACM international conference on Web search and data mining}}. ACM,
  \bibinfo{pages}{81--90}.
\newblock


\bibitem[\protect\citeauthoryear{Richardson, Dominowska, and Ragno}{Richardson
  et~al\mbox{.}}{2007}]%
        {richardson2007predicting}
\bibfield{author}{\bibinfo{person}{Matthew Richardson}, \bibinfo{person}{Ewa
  Dominowska}, {and} \bibinfo{person}{Robert Ragno}.}
  \bibinfo{year}{2007}\natexlab{}.
\newblock \showarticletitle{Predicting clicks: estimating the click-through
  rate for new ads}. In \bibinfo{booktitle}{\emph{Proceedings of the 16th
  international conference on World Wide Web}}. ACM, \bibinfo{pages}{521--530}.
\newblock


\bibitem[\protect\citeauthoryear{Shan, Hoens, Jiao, Wang, Yu, and Mao}{Shan
  et~al\mbox{.}}{2016}]%
        {shan2016deep}
\bibfield{author}{\bibinfo{person}{Ying Shan}, \bibinfo{person}{T~Ryan Hoens},
  \bibinfo{person}{Jian Jiao}, \bibinfo{person}{Haijing Wang},
  \bibinfo{person}{Dong Yu}, {and} \bibinfo{person}{JC Mao}.}
  \bibinfo{year}{2016}\natexlab{}.
\newblock \showarticletitle{Deep Crossing: Web-scale modeling without manually
  crafted combinatorial features}. In \bibinfo{booktitle}{\emph{Proceedings of
  the 22nd ACM SIGKDD International Conference on Knowledge Discovery and Data
  Mining}}. ACM, \bibinfo{pages}{255--262}.
\newblock


\bibitem[\protect\citeauthoryear{Thai-Nghe, Drumond, Horv{\'a}th, and
  Schmidt-Thieme}{Thai-Nghe et~al\mbox{.}}{2012}]%
        {thai2012using}
\bibfield{author}{\bibinfo{person}{Nguyen Thai-Nghe}, \bibinfo{person}{Lucas
  Drumond}, \bibinfo{person}{Tom{\'a}s Horv{\'a}th}, {and}
  \bibinfo{person}{Lars Schmidt-Thieme}.} \bibinfo{year}{2012}\natexlab{}.
\newblock \showarticletitle{Using factorization machines for student
  modeling.}. In \bibinfo{booktitle}{\emph{UMAP Workshops}}.
\newblock


\bibitem[\protect\citeauthoryear{Wang, Fu, Fu, and Wang}{Wang
  et~al\mbox{.}}{2017}]%
        {wang2017deep}
\bibfield{author}{\bibinfo{person}{Ruoxi Wang}, \bibinfo{person}{Bin Fu},
  \bibinfo{person}{Gang Fu}, {and} \bibinfo{person}{Mingliang Wang}.}
  \bibinfo{year}{2017}\natexlab{}.
\newblock \showarticletitle{Deep \& Cross Network for Ad Click Predictions}.
\newblock \bibinfo{journal}{\emph{arXiv preprint arXiv:1708.05123}}
  (\bibinfo{year}{2017}).
\newblock


\bibitem[\protect\citeauthoryear{Weinberger, Dasgupta, Langford, Smola, and
  Attenberg}{Weinberger et~al\mbox{.}}{2009}]%
        {weinberger2009feature}
\bibfield{author}{\bibinfo{person}{Kilian Weinberger}, \bibinfo{person}{Anirban
  Dasgupta}, \bibinfo{person}{John Langford}, \bibinfo{person}{Alex Smola},
  {and} \bibinfo{person}{Josh Attenberg}.} \bibinfo{year}{2009}\natexlab{}.
\newblock \showarticletitle{Feature hashing for large scale multitask
  learning}. In \bibinfo{booktitle}{\emph{Proceedings of the 26th Annual
  International Conference on Machine Learning}}. ACM,
  \bibinfo{pages}{1113--1120}.
\newblock


\bibitem[\protect\citeauthoryear{Zhang, Du, and Wang}{Zhang
  et~al\mbox{.}}{2016}]%
        {zhang2016deep}
\bibfield{author}{\bibinfo{person}{Weinan Zhang}, \bibinfo{person}{Tianming
  Du}, {and} \bibinfo{person}{Jun Wang}.} \bibinfo{year}{2016}\natexlab{}.
\newblock \showarticletitle{Deep learning over multi-field categorical data}.
  In \bibinfo{booktitle}{\emph{European conference on information retrieval}}.
  Springer, \bibinfo{pages}{45--57}.
\newblock


\end{thebibliography}

\end{document}